\documentclass{article}
\usepackage{ijcai23}

\usepackage{times}
\usepackage{soul}
\usepackage{url}
\usepackage[hidelinks]{hyperref}
\usepackage[utf8]{inputenc}
\usepackage[small]{caption}
\usepackage{graphicx}
\usepackage{amsmath}
\usepackage{amsthm}
\usepackage{booktabs}
\usepackage{algorithm}
\usepackage{algorithmic}
\usepackage[switch]{lineno}

\title{Automatic Concept Embedding Model (ACEM): \\No train-time concepts, No issue!}

\author{
    Rishabh Jain
    \affiliations
    University of Cambridge
    \emails
    rj412@cam.ac.uk
}

\begin{document}

\maketitle

\begin{abstract}
    Interpretability and explainability of neural networks is continuously increasing in importance, especially within safety-critical domains and to provide the social right to
    explanation. Concept based explanations align well with how humans reason, proving to be a good way to explain models. Concept Embedding Models (CEMs) are one such concept
    based explanation architectures. These have shown to overcome
    the trade-off between explainability and performance. However, they have a key limitation -- they require concept annotations for all their training data. For large datasets,
    this can be expensive and infeasible. Motivated by this, we propose Automatic Concept Embedding Models (ACEMs), which learn the concept annotations automatically.
\end{abstract}

\section{Introduction}
Many explainable AI (XAI) techniques aim to explain a models predictions by assigning importance to the model's features. However, various works have shown that these are unreliable
\cite{Adebayo-unreliable-XAI-2,Gimenez-unreliable-XAI-3,Ghorbani-unreliable-XAI-1} and do not increase human understanding or trust in the model
\cite{Kim-TCAV,Poursabzi-Manipulating-Model-Interpretability}. Kim et. al. \cite{Kim-TCAV} show that given identical feature-based explanations,
human subjects confidently find evidence for contradictory conclusions.

As a result, work has been done to obtain explanations in the form of high-level human understandable concepts \cite{Ciravegna-LEN,Ghorbani-ACE,Kim-TCAV,Koh-CBM}.
This is in line with how humans normally reason \cite{Zarlenga-CEM}. Methods like Testing with Concept Activation Vectors (TCAV) \cite{Kim-TCAV} act on black-box models
post-hoc and require hand-labelled examples of concepts. Providing these concept examples encompass bias and it might also not be feasible to do so due to the presence
of too many potential concepts.
Automatic Concept-based Explanations (ACE) \cite{Ghorbani-ACE} by Ghorbani et. al. offer a solution to this problem by automatically learning the concepts.

These previous methods focused only on obtaining global explanations. For various human-model interactions and human-in-the-loop settings, local explanations prove to be
vital. Concept Bottleneck Models (CBMs) \cite{Koh-CBM} by Koh, Nguyen and Tang et. al. is an architecture that learns to align models with the provided train-time concepts.
By dividing the model into separate concept encoding and label predicting parts, CBMs are able to provide both local and global explanations based on human-understandable concepts.
Concept Embedding Models (CEMs) \cite{Zarlenga-CEM} by Zarlenga and Barbiero et. al. bridge the gap between interpretability and task accuracy of CBMs when the concept
space is incomplete.

\begin{figure*}[tb]
    \centering
    \includegraphics[width=0.8\linewidth]{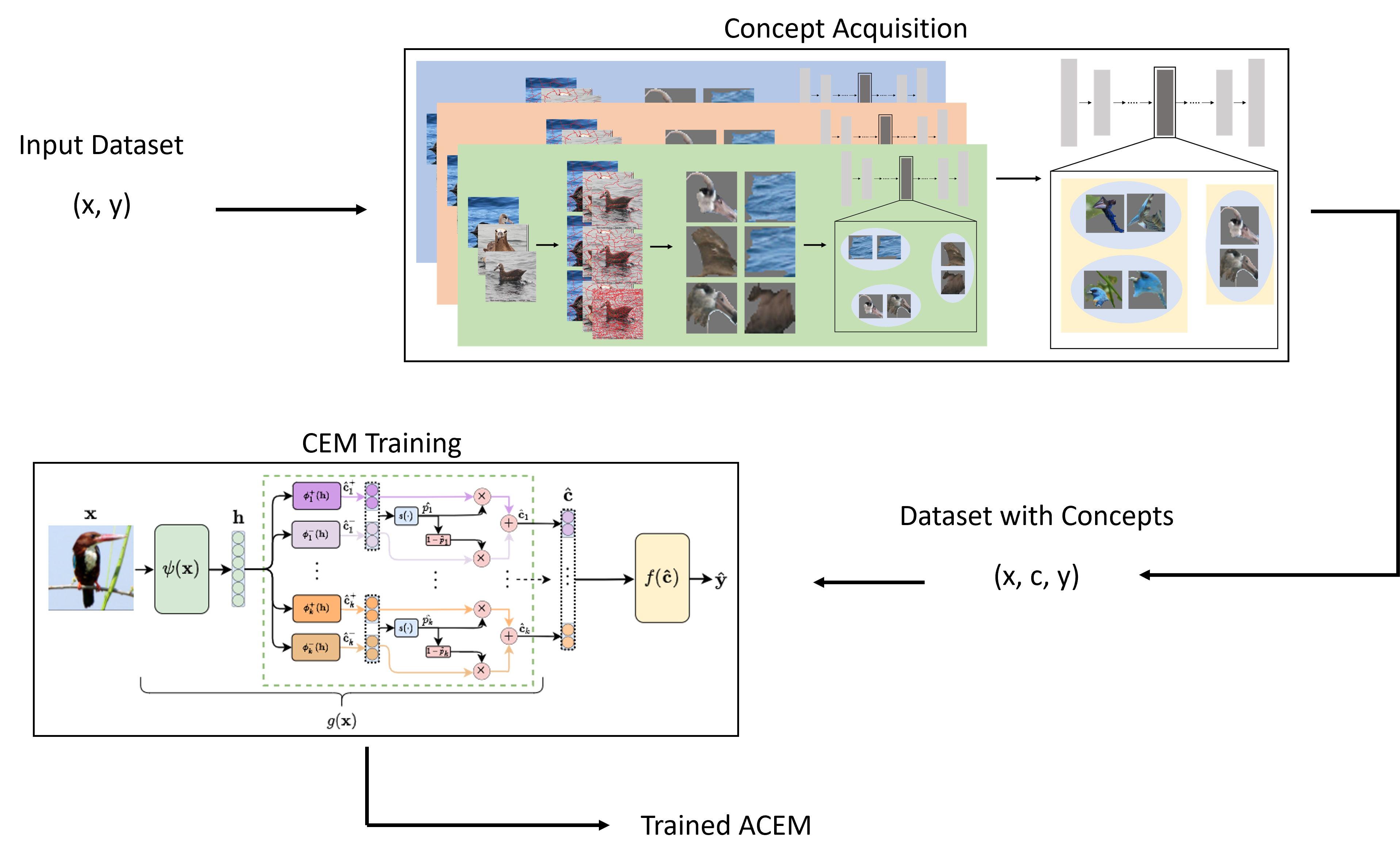}
    \caption{ACEM overview. The method takes a dataset without concept annotations and yields a trained ACEM model. This is done by first acquiring concepts for the dataset,
             using procedures inspired from ACE. Then the training dataset is annotated to allow traditional CEM training, yielding a CEM model which has been trained to use
             the acquired concepts for its task predictions. This model is dubbed as ACEM.}
    \label{fig:acem-overview}
\end{figure*}

In this paper, we relieve the concept labelling requirement of CEMs by combining the concept extraction of ACE with the model architecture and training strategies of
CEM. Doing so, we obtain Automatic Concept Embedding Models (ACEMs). These are models that do not require concept annotations at train-time, provide predictions based on
automatically discovered concepts and allows test time intervention on these concepts. Figure \ref{fig:acem-overview} shows an overview of the method proposed.

\section{Background}
\textbf{Concept Bottleneck Models} (CBMs) \cite{Koh-CBM} are a class of models that are `\emph{bottlenecked}' by an intermediate concept space $C$.
For the task of predicting the label $y \in Y$ given a sample $x \in X$, a CBM can be constructed using: (1) a concept encoding function $g : X \rightarrow C$
and (2) a label prediction function $f : C \rightarrow Y$.

When the concept space $C$ is not complete, there is a trade-off between performance and interpretability. \textbf{Concept Embedding Models} (CEMs) aim to overcome this
trade-off between performance and interpretability by representing concepts as a vector. The high-dimensional embedding vectors allow each concept to have extra supervised
learning capacity.

\textbf{Automatic Concept-based Explanations} (ACE) \cite{Ghorbani-ACE} is a post-hoc global explanation method, that does not require concept annotations/examples,
but rather learns them automatically. ACE segments images from a given class, and clusters these segments in the activation space of the final layers of a
state-of-the-art CNN trained on some large-scale dataset. These clusters form the concepts.

\section{Related Works}
This paper extends CBMs \cite{Koh-CBM} and CEMs \cite{Zarlenga-CEM} to use automatically generated concepts, with ACE \cite{Ghorbani-ACE} as the main technique for doing so.
Thus, these works are the most directly related to the paper.

Other works have attempted to extend CBMs to not require concept labels in different ways. Oikarinen et. al. \cite{Oikarinen-Label-Free-CBM} propose Label-Free CBM, which generates
the initial concept labels via GPT-3 \cite{Brown-GPT-3} using OpenAI APIs. Yuksekgonul et. al. \cite{Yuksekgonul-Post-Hoc-CBM} propose Post-Hoc Concept Bottleneck Models which
uses ConceptNet to generate the concepts.

\section{Automatic Concept Embedding Model}
In real-world settings, obtaining concept annotations is costly and rare. This is especially true for larger datasets containing millions of images.
Techniques like CBMs and CEMs require concept annotations for all training samples to actually train the model. On the other hand, techniques like
ACE do not require any concept annotations, but act on the model post-hoc and do not provide local explanations or allow interventions.

We propose Automatic Concept Embedding Model (ACEM) which aims to combine the bests of both worlds, so that we get models which do not require concept
annotations to train, provide local and global explanations and also support interventions. To do so, we divide the model training into two steps: (1) concept
acquisition, and (2) CEM training. Concept acquisition is done by using concepts found with ACE. These concepts are then used to annotate the training data for the CEM model.

\subsection{Concept Acquisition}
We decompose the concept acquisition step into two main steps -- (1) Concept generation for each class, and (2) Aggregating concepts across classes
to obtain a single concept space for each image.


\paragraph{Per-class concept generation} ACEM acquires concepts for each class by using the automatic concept generation of ACE.
ACE requires two inputs -- class examples and a trained model to explain. Using these, ACE segments the class images, clusters these segments into concepts and gives
importance for each concept using TCAV on the given model. However, for the importance calculation, here we do not have the trained model that we want to explain yet.
Thus, we here make the design choice to treat all detected concepts with equal importance. We investigate two design choices for the activation space for clustering --
`\emph{Pre-Trained Activation Space}' (PT$_{\text{act}}$) and `\emph{Black-Box Activation Space}' (BB$_{\text{act}}$).

In PT$_{\text{act}}$, we use Euclidean distance in the activation space of a final layer of GoogLeNet \cite{Szegedy-GoogLeNet} for calculating the perceptual similarity
between segments for clustering. This follows from the work of Zhang et. al. \cite{Zhang-Perceptual-Similarity}, and is also in line with what Ghorbani et. al. do for
ACE \cite{Ghorbani-ACE}.

Another approach would be to train a black-box model, without concept supervision, with similar architecture to our intended final CEM. Here we can use the output space
of the `concept encoding' part of the model (more appropriately the part that would be the concept encoding part if we had trained the model with concept supervision)
instead of the activation space of a layer of the GoogLeNet model. We dub this design choice as `\emph{Black-Box Activation Space}' (BB$_{\text{act}}$). Note that
in real-world settings, in order to get good generalisation, the model is going to be trained on some large-scale data set or the concept encoding part of the model
is going to consist of some pre-trained state-of-the-art general CNN (without some top layers), trained on some large-scale data set. Thus, this design choice also
makes sense for perceptual similarity.

\begin{figure*}[tb]
    \centering
    \includegraphics[width=0.7\linewidth]{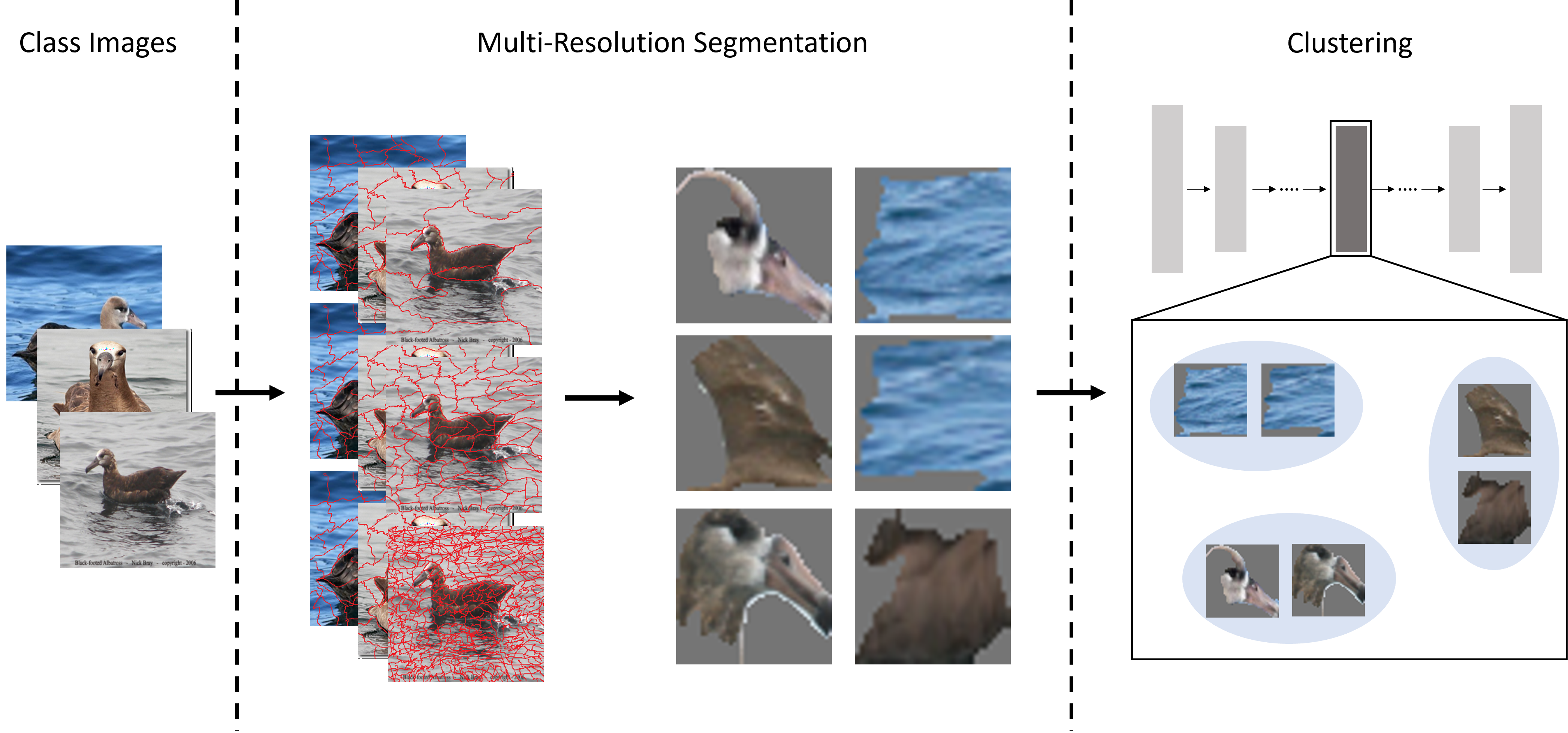}
    \caption{Concept Generation overview. Images of one class are taken as input and concepts for this class is generated. This is done by multi-resolution segmentation
             followed by clustering.}
    \label{fig:acem-concept-generation}
\end{figure*}

ACEM starts with the given dataset of images and their class labels. To obtain the concepts for some specific class, first all the images are segmented with
multiple-resolutions, using a segmentation strategy.
We use SLIC \cite{Achanta-SLIC} with three different resolutions, similar to Ghorbani et. al. \cite{Ghorbani-ACE}. These segments
are clustered together using a clustering strategy (we use k-Means), with similarity score determined using the design choice for the activation space, followed by outlier filtering.
Due to absence of the model, no saliency computation is done. These generated
concepts are then passed to the concept aggregation step. Figure \ref{fig:acem-concept-generation} shows an overview of concept generation.

\paragraph{Concept Aggregation}
Per-class concept generation gave us a set of concepts for each class, with each concept being simply a cluster of rescaled patches. We take each of these concepts to be describing a
hypersphere within the activation space we chose, with the centre as the centroid of the related rescaled patches, and radius as the maximum distance between this centre and the
rescaled patches.

Concepts might be shared across different classes. To combine these, we run a concept aggregation step, clustering using k-Means,
on the concept centres for all the concepts of all the classes. The number of clusters here is chosen to match the desired concept space size.

\subsection{CEM training}
Training CEMs requires a dataset with each element in $X \times C \times Y$, where $X$ is the input space, $C$ is the concept space and $Y$ is the output space.
ACEMs start with a dataset with each element in $X \times Y$. Thus, to train the CEM appropriately, we need to first compute the concept labels for each data point.

In the concept acquisition phase, we generated a concept space $C$, and found concept definitions for each concept (the concept hyperspheres). We label the training
data by segmenting the images, and checking whether these segments lie in the hypersphere described by the concepts of the class of the image.
The union of satisfied concepts is taken to be the concept labelling for the image.

This leads to a dataset of the form $X \times C \times Y$, which we can use to train a CEM model are required by \cite{Zarlenga-CEM}.
This final model will learn to make predictions using the concepts as obtained from the concept acquisition phase.

\section{Experiments}
We evaluate the performance of ACEM on two datasets -- (1) MNIST-ADD \cite{LeCun-MNIST}, and (2) Caltech-UCSD Birds (CUB) dataset \cite{Wah-CUB}.
We compare its performance with that of CEM, CBM, Hybrid-CBM and a black-box model. In order to keep the comparison fair, we choose each model with hyperparameters and architecture
choices as similar to the ACEM model's architecture as possible.

We evaluate the task accuracy and concept alignment scores (CAS), as used by Zarlenga and Barbiero et. al. \shortcite{Zarlenga-CEM}.
CAS aims to measure the how much the learnt concept representations can be trusted as faithful representations of the ground truth concept labels.
A higher CAS means a more aligned concept representation.

\subsection{MNIST-ADD Dataset}
\label{sec:exp-mnist-add}
\begin{figure}[tbhp]
    \centering
    \includegraphics[width=1.0\linewidth]{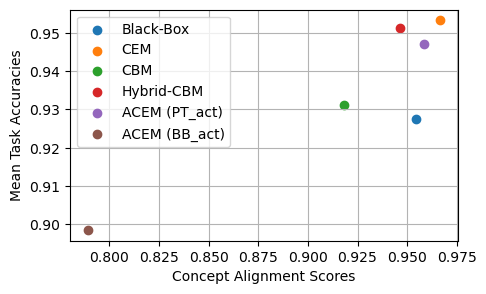}
    \caption{Mean test task accuracies versus the CAS values for the different strategies on the MNIST-ADD dataset.}
    \label{fig:mnist-add-eval}
\end{figure}

Each sample in the dataset consists of an image formed by concatenating two images, each being a handwritten image of a number from 0 to 5. The desired task is to predict the
sum of the numbers represented by the concatenated images. The ground-truth concepts are taken to be the concatenated one hot representation of the number represented by each image.
Thus, this dataset consists of 12 complete concept annotations with 11 task labels. Here we use a simple CNN for the concept encoding part of the models.
We plot both the mean test task accuracies and CAS for the different strategies in Figure \ref{fig:mnist-add-eval}.

We observe that  PT$_{\text{act}}$ outperforms BB$_{\text{act}}$ in terms of both task accuracy and CAS.
This is not too surprising here because the concept encoder for the black box model in BB$_{\text{act}}$ is a very basic
CNN trained on a very limited task. Thus, we cannot expect the results of Zhang et. al. \shortcite{Zhang-Perceptual-Similarity} for perceptual similarity to hold here.

We see that, as expected, CEM outperforms all of the other
models in terms of both task accuracies and CAS. Notably, ACEM (PT$_{\text{act}}$) outperforms the black-box and CBM models in terms of both the task accuracies and CAS.
This means that ACEM (PT$_{\text{act}}$) is able to learn useful concepts which align well with the ground truth labels. 

Thus, we can see a clear benefit of using ACEM over a black-box model here when no concept annotations are available. In addition, since the base architecture is still a CEM,
we can perform concept interventions easily.

\subsection{CUB Dataset}
\begin{figure}[tbhp]
    \centering
    \includegraphics[width=1.0\linewidth]{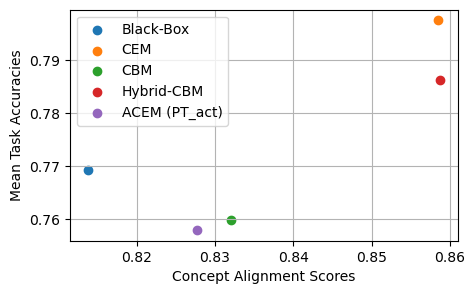}
    \caption{Test task accuracies versus the CAS values for the different strategies on the CUB dataset.}
    \label{fig:cub-eval}
\end{figure}

The task here is to predict the species of a bird, given its image. The dataset also contains complete annotations for the features of the bird visible in each image. The dataset consists
of 112 complete concept annotations with 200 task labels.
We plot both the test task accuracies and CAS for the different strategies in Figure \ref{fig:cub-eval}.



Here we see that the use of ACEM (PT$_{\text{act}}$) over a black-box model leads to better concept alignment, but poorer test task accuracy. This contrasts with
MNIST-ADD dataset, where we see a clear improvement in both. One reason might be that the images in CUB dataset consist of birds in the nature. Thus, they consist of
various backgrounds like forests and seas. This means that segmentation followed by clustering is likely to lead to concepts that are focused on these background
features. Thus, our learnt concept space consists of some unrelated concepts, leading to poorer test task accuracies.

One way to prevent this would be to have some form of concept filtering during the concept acquisition phase. This could be done in the concept aggregation step,
where we can consider the class frequencies and concept sizes to decide which concepts seem to be irrelevant background features, and so can be ignored.
Another potential way to solve this is to train a black-box model of similar architecture to the desired CEM, and calculate saliency for each concept, potentially using
TCAVs as done by Ghorbani et. al. \cite{Ghorbani-ACE} in ACE. This might be especially lucrative if we are using BB$_{\text{act}}$, which also requires training
of the black-box model.

\section{Summary \& Conclusion}
In this paper, we successfully combined ACE \cite{Ghorbani-ACE} with CEMs \cite{Zarlenga-CEM} to obtain ACEMs. This led to a model that does not require concept
annotations, even during training time, just like ACE, and is capable of giving explanations using `human-understandable' concepts and allows for concept interventions,
just like CEMs. We then showed, using MNIST-ADD dataset, that this ACEM model gave better test task performance than a similarly shaped black-box model, while also
being more aligned with the ground-truth concepts. Furthermore, its overall performance was close to CEMs, despite not requiring
any train-time concept annotations.

We then try this approach on a more complex and real-world dataset, CUB. Here we see that ACEM's test task performance was poorer than the black-box model, although
the concept alignment was still better. The depreciation in the test task accuracy by using ACEMs over the black-box model is not too great, and so we can potentially
afford to take that hit, and obtain a more interpretable model which allows for intervention. Further work is needed to bridge this gap.

To conclude, ACEMs allow us to get an interpretable model that allows for concept interventions in settings where obtaining concept annotations would not be feasible, without
taking a (too great) hit to the task performance. These could be large scale datasets, like ImageNet \cite{Deng-ImageNet} and Places \cite{Zhou-Places}, which
contain millions of data samples.

\appendix

\section*{Ethical Statement}
There are no ethical issues.

\bibliographystyle{named}
\bibliography{ijcai23}

\clearpage

\section{Sample ACEM predictions}

\begin{figure*}[tb]
    \centering
    \includegraphics[width=0.8\linewidth]{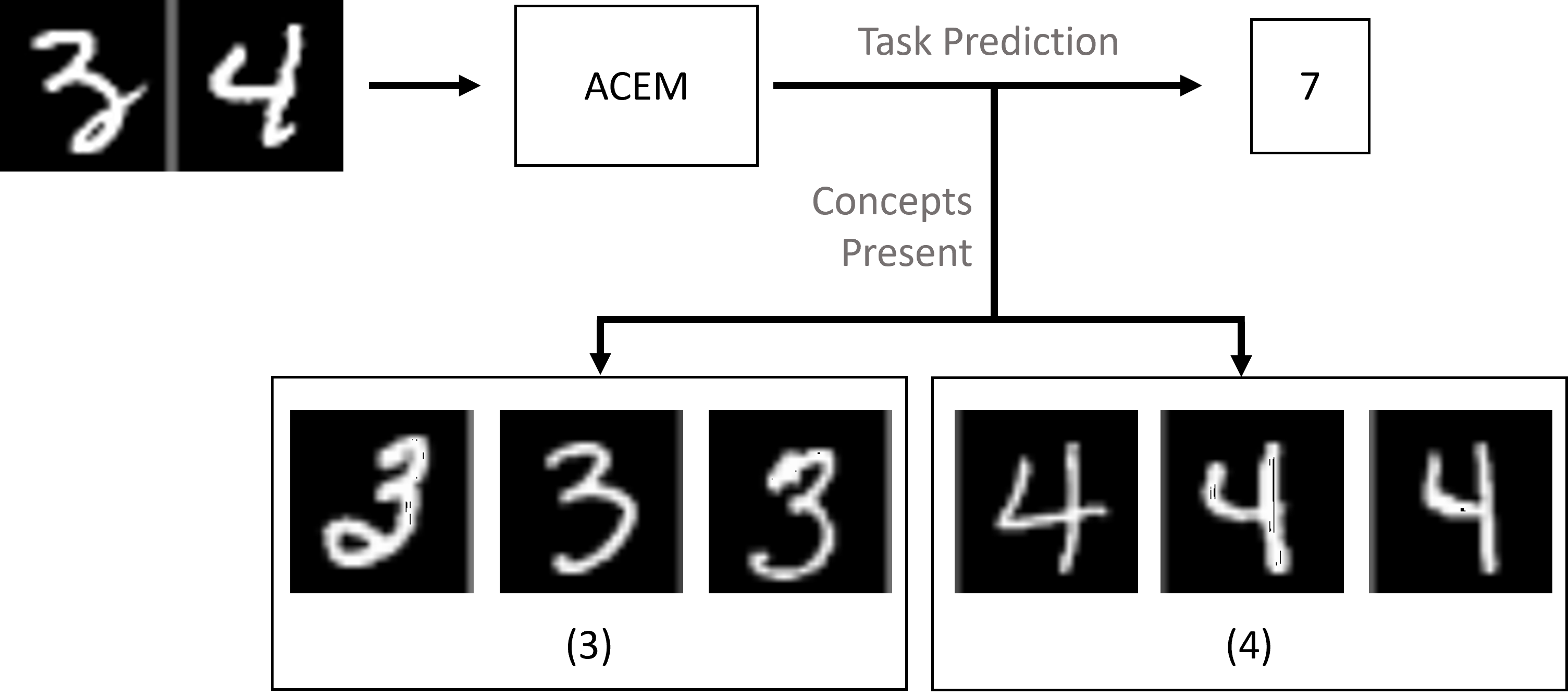}
    \caption{Example ACEM prediction of a test sample from MNIST-ADD dataset. The sum is correctly predicted as `7'. The figure also shows the concepts used in the form
             of some sample images of the concept, with a potential reasoning as to what the concepts could represent in brackets below the concept.}
    \label{fig:mnist-add-sample}
\end{figure*}

\begin{figure*}[tb]
    \centering
    \includegraphics[width=0.8\linewidth]{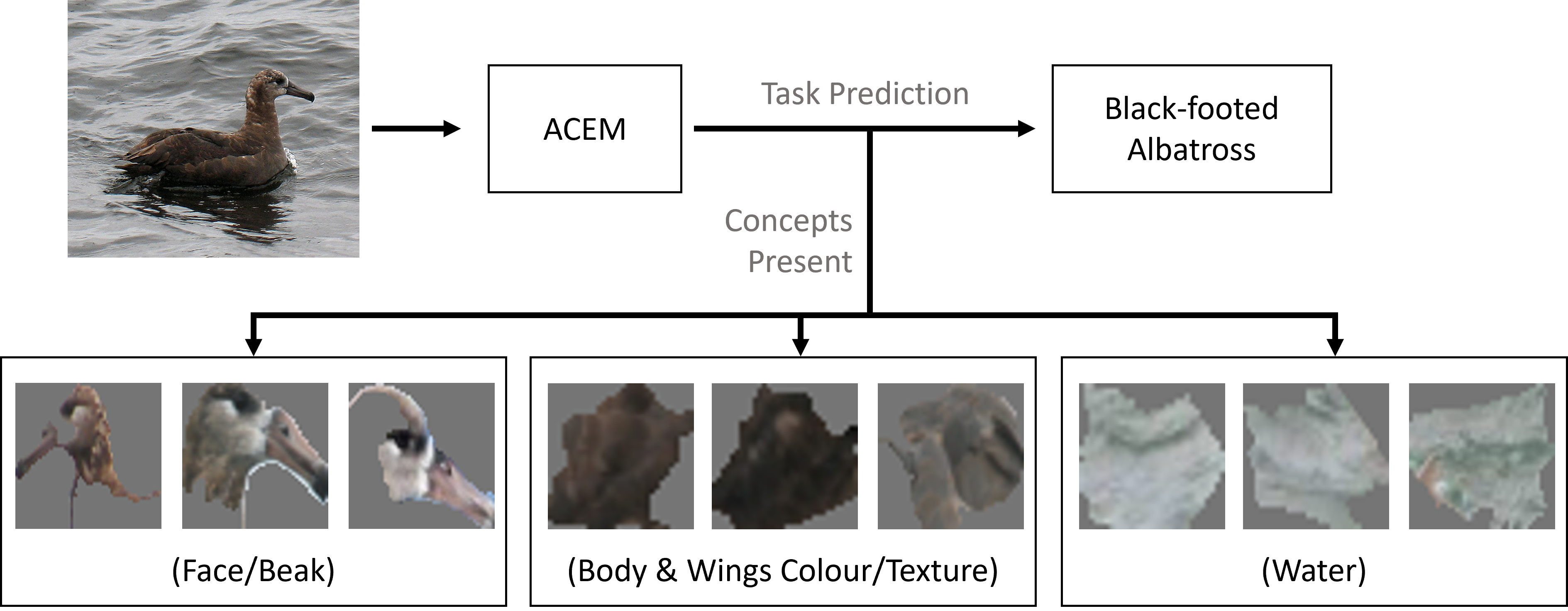}
    \caption{Example ACEM prediction of a test sample from CUB dataset. The sum is correctly predicted as `Black-footed Albatross'. The figure also shows the concepts used in the form
             of some sample images of the concept, with a potential reasoning as to what the concepts could represent in brackets below the concept.}
    \label{fig:cub-sample}
\end{figure*}

Figure \ref{fig:mnist-add-sample} shows an example of ACEM prediction with a sample from MNIST-ADD dataset. The input is a concatenation of numbers `3' and `4'. The sum is correctly
predicted to be `7'. We also show the concepts present, as predicted by ACEM. The concepts are represented by a collection of segments, as determined in the concept acquisition phase.
It is easy to see that the concept on the left of the image represents the number `3', while the concept on the right represents number `4'. These are indeed the concepts that should
be present.

Figure \ref{fig:cub-sample} shows an example of ACEM prediction with a sample from CUB dataset. The bird is indeed correctly predicted to be a `Black-footed Albatross'.
We also show the concepts present as determined by ACEM. Below each image, we note what they seem to be representing (since ACEM requires no concept annotations, the concepts
discovered are unlabelled). Each of these concepts are sensible in one way or other. This is because the face and beak of the bird, as well as the body and plumage of the bird
are good indicators for what species a bird might be. For example, Black-footed Albatross have markings below their eyes \cite{Floyd-Black-Footed}, and indeed we see
that in each image of the face concept here. The final concept represents water. Although this might not directly correlate with the species, it is still somewhat indicative of the
species here. This is because Black-footed Albatross are a species of seabirds, and thus are expected to be found with a sea in the background.

\section{ACEM Ablation Experiment}

In the paper, we use ACE with CEM. However, we can also use ACE with other models, like CBM and Hybrid-CBM. In this experiment, we compare the performance of these choices
on the MNIST-ADD dataset. Table \ref{table:mnist-add-task-accuracies} shows the results from this experiment.

\begin{table}[tbh]
    \centering
    \begin{tabular}{|c|c|}
        \hline
        \bfseries Method & \bfseries Task Accuracy \\
        \hline
        Black-Box & $0.928 \pm 0.003$ \\
        \hline
        CEM & $0.953 \pm 0.002$ \\
        \hline
        CBM & $0.931 \pm 0.002$ \\
        \hline
        Hybrid-CBM & $0.951 \pm 0.003$ \\
        \hline
        ACEM (PT$_{\text{act}}$) & $0.947 \pm 0.001$ \\
        \hline
        ACE-CBM (PT$_{\text{act}}$) & $0.735 \pm 0.0005$ \\
        \hline
        ACE-Hybrid-CBM (PT$_{\text{act}}$) & $0.920 \pm 0.001$ \\
        \hline
        ACEM (BB$_{\text{act}}$) & $0.898 \pm 0.005$ \\
        \hline
        ACE-CBM (BB$_{\text{act}}$) & $0.781 \pm 0.0002$ \\
        \hline
        ACE-Hybrid-CBM (BB$_{\text{act}}$) & $0.890 \pm 0.006$ \\
        \hline
    \end{tabular}
    \caption{Mean $\pm$ Standard Deviation of Test Task Accuracies for the different strategies, as calculated over 3 runs, on the MNIST-ADD dataset.}
    \label{table:mnist-add-task-accuracies}
\end{table}

We find that ACEM outperforms both ACE-CBM and ACE-Hybrid-CBM. This means that the extra-supervised capacity of CEMs does help ACEMs and is needed to ensure a high
performance.

\section{Experiment Tables}

In the experiments section, we present the tables for the different experiments. Here we note the tables for the corresponding figures. Table \ref{table:mnist-add-task-accuracies}
shows the test task accuracies on the MNIST-ADD dataset, while Table \ref{table:mnist-add-cas} shows the CAS scores for the models on the MNIST-ADD dataset.

Table \ref{table:cub-task-accuracies} and Table \ref{table:cub-cas} show the test task accuracies and the CAS scores for the models on the CUB dataset, respectively.

\begin{table}[tbhp]
    \centering
    \begin{tabular}{|c|c|}
        \hline
        \bfseries Method & \bfseries CAS \\
        \hline
        Black-Box & 0.95 \\
        \hline
        CEM & 0.97 \\
        \hline
        CBM & 0.92 \\
        \hline
        Hybrid-CBM & 0.95 \\
        \hline
        ACEM (PT$_{\text{act}}$) & 0.96 \\
        \hline
        ACEM (BB$_{\text{act}}$) & 0.79 \\
        \hline
    \end{tabular}
    \caption{Concept Alignment Scores for the different strategies on the MNIST-ADD dataset.}
    \label{table:mnist-add-cas}
\end{table}

\begin{table}[tbhp]
    \centering
    \begin{tabular}{|c|c|}
        \hline
        \bfseries Method & \bfseries Task Accuracy \\
        \hline
        Black-Box & 0.77 \\
        \hline
        CEM & 0.80 \\
        \hline
        CBM & 0.76 \\
        \hline
        Hybrid-CBM & 0.79 \\
        \hline
        ACEM (PT$_{\text{act}}$) & 0.76 \\
        \hline
    \end{tabular}
    \caption{Test Task Accuracies for the different strategies on the CUB dataset.}
    \label{table:cub-task-accuracies}
\end{table}

\begin{table}[tbhp]
    \centering
    \begin{tabular}{|c|c|}
        \hline
        \bfseries Method & \bfseries CAS \\
        \hline
        Black-Box & 0.81 \\
        \hline
        CEM & 0.86 \\
        \hline
        CBM & 0.83 \\
        \hline
        Hybrid-CBM & 0.86 \\
        \hline
        ACEM (PT$_{\text{act}}$) & 0.83 \\
        \hline
    \end{tabular}
    \caption{Concept Alignment Scores for the different strategies on the CUB dataset. Computing this for the whole test split of the dataset is expensive.
             We approximate this by batching the calculation and calculating this for only a subset of the concepts. Thus, these might not lie in the confidence
             intervals as calculated by Zarlenga and Barbiero et. al.}
    \label{table:cub-cas}
\end{table}

\end{document}